\documentclass[a4paper]{article}

\usepackage{INTERSPEECH2022}

\usepackage{amsmath,graphicx}
\usepackage{multicol,multirow,array}

\usepackage{xcolor}

\definecolor{darkspringgreen}{rgb}{0.09, 0.7, 0.27}

\title{Streaming Align-Refine for Non-autoregressive Deliberation}
\name{Weiran Wang $\qquad$ Ke Hu $\qquad$ Tara N. Sainath}
\address{Google, Inc.}
\email{\{weiranwang, huk, tsainath\}@google.com}

\begin{document}
\maketitle
\begin{abstract}
We propose a streaming non-autoregressive (non-AR) decoding algorithm to deliberate the hypothesis alignment of a streaming RNN-T model. Our algorithm facilitates a simple greedy decoding procedure, and at the same time is capable of producing the decoding result at each frame with limited right context, thus enjoying both high efficiency and low latency. These advantages are achieved by converting the offline Align-Refine algorithm to be streaming-compatible, with a novel transformer decoder architecture that performs local self-attentions for both text and audio, and a time-aligned cross-attention at each layer.
Furthermore, we perform discriminative training of our model with the minimum word error rate (MWER) criterion, which has not been done in the non-AR decoding literature.
Experiments on voice search datasets and Librispeech show that with reasonable right context, our streaming model performs as well as the offline counterpart, and discriminative training leads to further WER gain when the first-pass model has small capacity.
\end{abstract}
\section{Introduction}
\label{sec:intro}
\vspace*{-.5ex}

% Deliberation is a two-pass modeling paradigm where a second-pass model is employed to refine decoding results of a first-pass model~\cite{xia2017deliberation}, by either re-decoding or rescoring first-pass hypotheses.
% Both first-pass hypotheses and ground truth pairs are presented in a supervised way for training the second-pass model.
% Previously, second-pass LAS models based on long-short term memory (LSTM) networks~\cite{hu2020deliberation} or transformer decoder~\cite{hu2021transformer} were shown to significantly improve Google voice search quality over the first-pass and an acoustic-rescoring model~\cite{sainath2020streaming}.
% One potential disadvantage of prior deliberation methods, however, is that the second-pass decoder still works in an autoregressive manner which can be slow.

There has been a % recent 
surge of interest in non-autoregressive (non-AR) automatic speech recognition (ASR) models that are not constrained to decode in a left-to-right manner~\cite{ghazvininejad2019MaskPredict,chen2019listen,higuchi2020mask,higuchi2021improved,deng2022improving,wang2021streaming,chan20,chi2020align}.
These models make parallel update steps during inference, i.e., each decoding step can modify multiple or all positions of previous step results simultaneously.
And they are attractive due to the simplicity and efficiency of their inference procedure.

Non-AR models can be largely categorized into two classes. The first class of models iteratively refine the label sequences~\cite{chen2019listen,higuchi2020mask,higuchi2021improved,wang2021streaming}, following the general framework of mask-predict~\cite{ghazvininejad2019MaskPredict}: in each refinement step, certain positions of the input label sequence % (e.g., the least confident predictions) 
are replaced by a special \texttt{[mask]} token, and the model learns to predicts all masked tokens simultaneously given the partially observed token sequence. A challenge to this approach is to estimate the length of the ground truth label sequence, for which heuristics based on CTC decoding results~\cite{higuchi2020mask,deng2022improving} and the dynamic length prediction task~\cite{higuchi2021improved} have been developed.

The second class of non-AR methods perform iterative refinement instead on alignments, which are sequences containing underlying tokens (including blanks for non-emission) at each frame. 
Representative methods in this class are Imputer~\cite{chan20} and Align-Refine~\cite{chi2020align,wang2022alignrefine}. The former 
gradually reveals positions of a fully masked alignment in a fixed number of steps, and its training procedure suffers from exposure bias. The latter updates complete alignment in each refinement step, effective allowing complex edits to the label sequences. 
% The CTC loss~\cite{graves2006connectionist} and its variants play a pivotal role in both methods, and they are the fundamental reason why parallel decoding can work well.

Previous work~\cite{wang2022alignrefine} has proposed a practical use case of Align-Refine for deliberation~\cite{xia2017deliberation,hu2020deliberation}. The authors use a small causal RNN-T~\cite{Graves_12a,He_18a}, which runs fast and has reasonable accuracy, to generate the initial hypothesis alignment, as opposed to using CTC~\cite{graves2006connectionist} as first-pass in the original algorithm of~\cite{chi2020align}.
% \KH{Maybe also mention how deliberation idea is incorporated here.} 
The authors then apply Align-Refine to the initial alignment for a few steps to generate new hypotheses of improved accuracy.

We note however, an important aspect of ASR---latency---has not been addressed by existing alignment-based non-AR methods.
In this work, we develop a streaming version of Align-Refine, 
which is capable of producing results through greedy decoding as data comes in, with controllable delay at each frame; this allows us to extend the use of Align-Refine into application scenarios with more stringent requirements on latency.
To achieve this goal, we propose a novel transformer decoder architecture that performs local self-attentions for both text and audio separately, and a time-aligned cross-attention at each layer; this architecture incorporates audio right context without incurring unnecessary model delay. While there have been use of other non-AR methods in the streaming mode~\cite{wang2021streaming,fujita2021toward}, the inference procedures of these methods are not as simple%\KH{(efficient?)} 
as ours. 

Furthermore, given our clean formulation and inference procedure, we perform discriminative training of streaming Align-Refine, with the minimum word error rate (MWER) criterion~\cite{prabhavalkar2018}. We found discriminative training to provide further WER gain when the first-pass model has small capacity. To the best of our knowledge, this is the first time sequence training is used in the non-AR decoding literature.
Experimental results on voice search datasets and Librispeech show that with reasonable right context, streaming Align-Refine performs similarly well as the offline counterpart, and compares favorably against existing non-AR methods and deliberation methods.

\section{Streaming Non-AR deliberation}
\label{sec:method}

\subsection{Review of offline Align-Refine}
\label{sec:rnnt}
\vspace*{-.5ex}

We generally follow the framework of~\cite{wang2022alignrefine} to apply Align-Refine~\cite{chi2020align} for deliberation. We use a streaming RNN-T~\cite{Graves_12a, He_18a} to generate first-pass hypothesis by autoregressive beam search, which achieves both good accuracy (by modeling label dependency) and low latency (without using right context).
% The RNN-T model has a causal encoder $enc_0$ that extracts audio features from the input utterances $X$, denoted by $enc_0(X)$ with length (number of frames) $T'$. The decoder, % of RNN-T, 
% denoted by $dec_0$, consists of the prediction network for modeling label dependency similarly to a language model, and the joint networks for combining audio and language model features and outputting per-frame posterior of labels.
Beam search returns the \emph{alignment} of the most probable hypothesis for each input utterance. The hypothesis alignment is a sequence of discrete tokens corresponding to the inferred labels of each frame; each label can be \texttt{<blank>} to indicate non-emission.
% Denote the alignment of consideration (e.g., the alignment for the 1-best hypothesis) from RNN-T by $A^0 (X)$ with length $T$. Note $T$ may be greater than $T'$, as RNN-T can output multiple labels at a frame.

\begin{figure}[t]
    \centering
    \includegraphics[width=0.85\linewidth, page=1, bb=50 10 600 400]{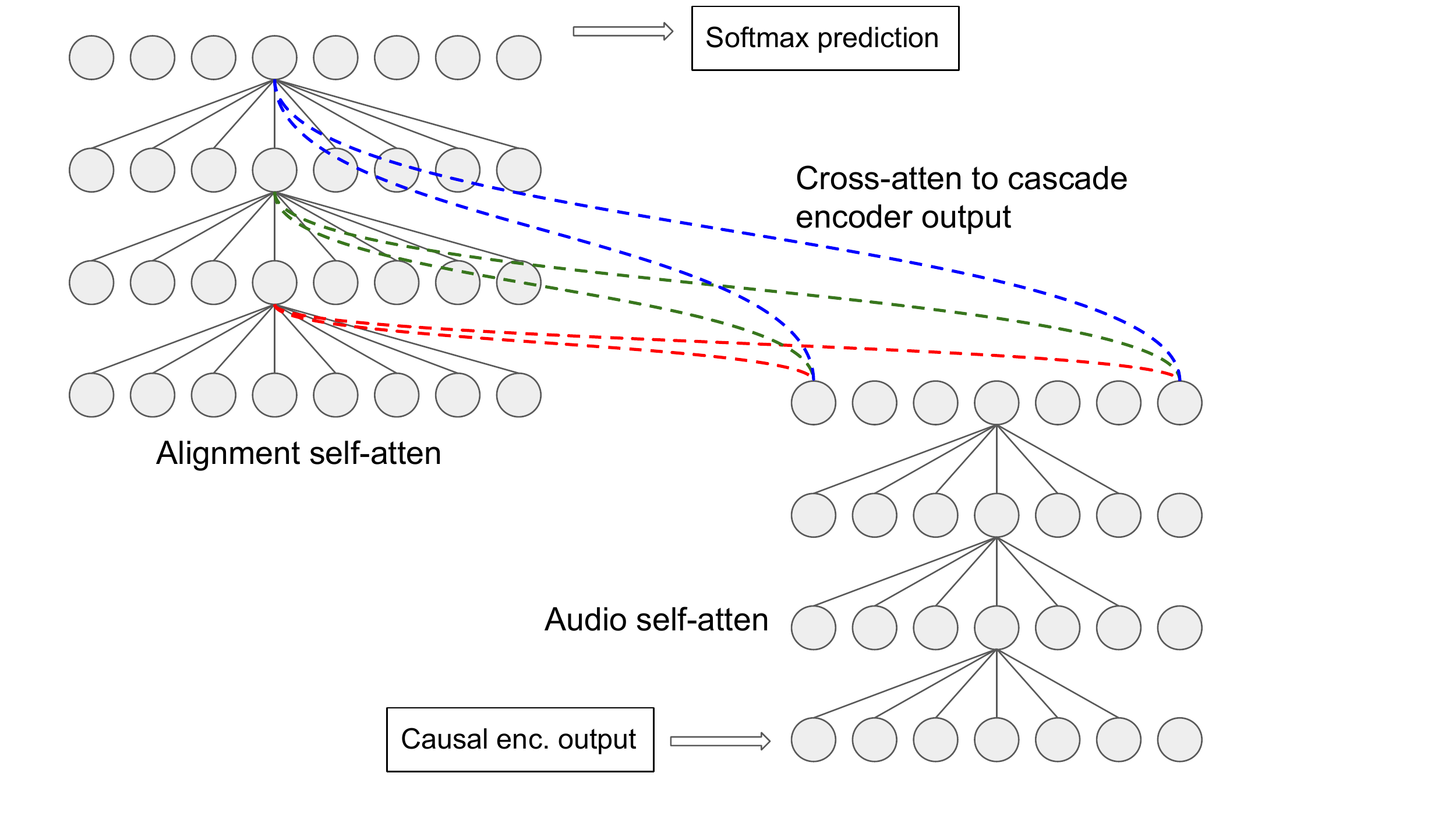} 
    \vspace*{-2ex}
    \caption{Transformer architecture of offline Align-Refine~\cite{wang2022alignrefine}.%\KH{in general I feel Fig. 1 can be removed since Fig. 2 is the new thing.}.
    }
    \label{fig:offline_alignrefine}
    \vspace*{-2ex}
\end{figure}

The first-pass hypothesis alignment is fed to the Align-Refine decoder for $S$ steps of iterative refinement. Each refinement step takes in an initial alignment and outputs a complete updated alignment; all steps share the same model parameters.
Similarly to the decoder module in attention-based model~\cite{Vaswani_17a}, the Align-Refine decoder consists of a series of transformer layers to integrate the text-side information from alignment and the audio-side information from encoder output.
A schematic diagram of the transformer architecture of~\cite{wang2022alignrefine} is provided in Figure~\ref{fig:offline_alignrefine}.
On the text side ("Alignment self-atten"), each transformer layer performs a self-attention for the text features, and use the result as query to perform cross-attention on audio features, %\KH{mention Fig. 1?}
whose result is in turn used as text-side features for the next layer. A final softmax layer is used on top of the last transformer layer output to predict labels and \texttt{<blank>}'s.
On the audio side ("Audio self-atten"), the authors propose to use additional non-causal cascaded encoder on top of the causal encoder of first-pass RNN-T, to extract audio features of rich right context; this architectural design indeed significantly boost the deliberation accuracy. Since the model of~\cite{wang2022alignrefine} works in the offline mode, the alignment self-attentions use full context, and each transformer layer attends to the cascaded encoder output with full context.
All layers are trained jointly using the CTC loss~\cite{graves2006connectionist}, which marginalizes alignments between CTC predictions and ground truth label sequences.

The CTC greedy alignment, obtained by picking the most probably token for each frame in parallel, is used as the output alignment and fed to the next refinement step. The greedy alignment after all $S$ steps is collapsed (by removing repetitions and then \texttt{<blank>}'s) into a label sequence as the final decoding result. The overall maximum likelihood estimation (MLE) loss % for Align-Refine 
is the averaged CTC losses at all steps: for input utterance $X$ with label sequence $y$, we minimize $\ell_{\text{MLE}}^S (X, y):= - \frac{1}{S} \sum_{i=1}^{S} \log P_{ctc}^i (y|X)$ where $\log P_{ctc}^i (y|X)$ is the full-sum conditional log-probability under the CTC model at step $i$. % (note we minimize $- \log P_{ctc} (y|X)$ in maximum likelihood training for CTC).

% \TS{this first sentence can probably go when you discuss RNN-T in the beginning of the second, it doesnt flow with the rest of the para/previos para} 
%In~\cite{wang2022alignrefine}, the authors propose to use additional non-causal cascaded encoder on top of the causal encoder of first-pass RNN-T, to extract audio features of rich right context; this architectural design indeed significantly boost the deliberation accuracy. 

% \TS{i'd reference the figure before you give the description in the previous para to make sure the reader does not get lost}

%  the refinement steps potentially allow complicated edits (insertion, deletion, substitution) from initial RNN-T hypotheses.

%     Second, the regular attention decoder trains and decodes in the autoregressive fashion and though only past label history is considered for predicting the current label, whereas the Align-Refine decoder employs full-context attention to capture also ``future" label information. Therefore, rich label context is already built into the model predictions and parallel decoding works well, alleviating us from repeatedly conditioning on the past as in autoregressive decoding.

\subsection{Streaming Align-Refine}
\label{sec:improvements}
\vspace*{-1ex}

We make architectural changes to convert Align-Refine to be streaming compatible. % and to minimize the model delay. 
First, we ensure the alignment self-attention is "local", so that the attention context vector of each frame only depends on a local window around itself, i.e., representation of each token is computed by attending to a number of previous tokens and a small number of future tokens. The total amount of right context accumulates with the depth of the architecture and becomes its model delay, e.g., if we have $L=6$ layers with per-layer right context $C=2$, the model waits for $L\times C=12$ future frames to output for the current frame.

Second, we ensure the cross-attention between alignment audio features is "local", so that the context vector of alignment frame (acting as the "query") only depends a window of audio feature frames (acting as the "key" and "value"). Note that owing to the RNN-T alignment topology~\cite{Graves_12a}, the alignment sequence and audio sequence are of different lengths. As an example, assume the input utterance has transcription "\texttt{hello world}" which is decomposed into $3$ wordpieces \{\texttt{\_hello}, \texttt{\_wor}, \texttt{ld}\}, and assume the encoder output consists of 5 audio frames. Then a plausible RNN-T alignment is
\begin{align*}
\texttt{<b> \_hello <b> <b> <b> \_wor ld <b>}
\end{align*}
where \texttt{<b>} denotes \texttt{<blank>}. Each \texttt{<blank>} token advances the audio frame index by $1$ whereas each non-blank token indicates label emission without advancing in time. By counting the number of previous blanks, we determine the "time", or equivalently the audio frame index, when each token is emitted. For the above example, tokens are output at frame indices (0-based)
% \KH{I think the below can be removed given there is a bit of space issue going on. The above example is good. But it is up to you, Weiran.}
\begin{align*}
\texttt{\qquad\quad 0\qquad 1\qquad 1\qquad 2\qquad 3\qquad 4\qquad 4\qquad 4\qquad}
\end{align*}
where \texttt{\_wor} and \texttt{ld} are both emitted at audio frame 4 as allowed by RNN-T.
We use these timestamp information to construct "local" cross-attention, so that a token emitted at audio frame index $t$ attends to a window of audio frames around $t$.

\begin{figure*}[t]
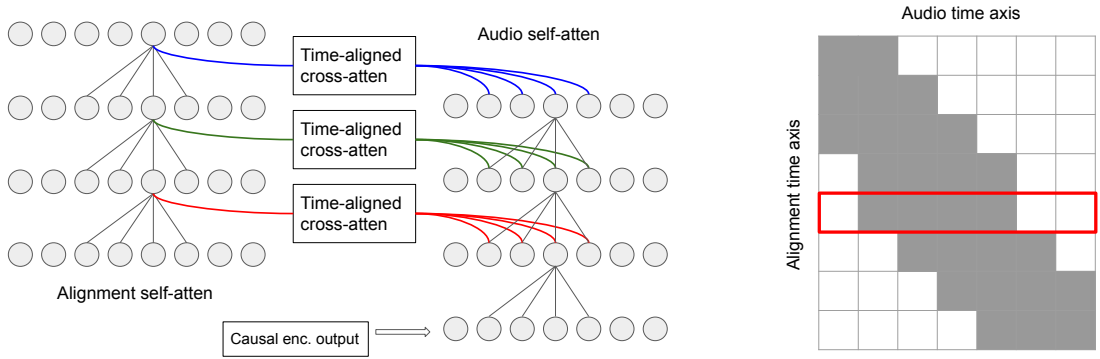

    \centering
    \begin{tabular}{@{}c@{\hspace{0.08\linewidth}}c@{}}
    \includegraphics[width=0.52\linewidth, page=2, bb=25 15 635 350, clip]{streaming_align_refine_figs.pdf} &
    \includegraphics[width=0.31\linewidth, page=3, bb=175 30 540 380, clip]{streaming_align_refine_figs.pdf} 
    \end{tabular}
    \vspace*{-3ex}
    \caption{Left: Transformer architecture of streaming Align-Refine where audio self-attention is synchronized with alignment self-attention. 
    % We omit the softmax layer on top of transformer architecture to avoid cluttering. 
    Right: The attention mask with shape $8\times 7$ shared by all time-aligned cross-attention operations in the left figure. Black indicates non-zero attention weights. Highlighted row indicates that frame 4 of the alignment feature sequence attends to a window around frame 3 of the audio feature sequence, with a left context of 2 frames and a right context of 1 frame.}
    \label{fig:streaming_alignrefine}
    \vspace*{-1.5ex}
\end{figure*}

Third, we propose a novel decoder architecture, shown in Figure~\ref{fig:streaming_alignrefine}, to utilize additional audio self-attention like cascaded encoder but without incurring further delay.
% If we follow the scheme in Figure~\ref{fig:offline_alignrefine} to first compute audio self-attention by a cascaded encoder with certain right context, and then compute the cross-attention between alignment features and cascaded encoder output, the right context from both modules contribute to the final model delay. To avoid this issue, we propose a novel transformer architecture that synergistically combines the two modules into one. 
For each layer, besides the alignment self-attention and time-aligned cross-attention, we compute in parallel an audio self-attention with the same amount of right context in "time" (audio frame index). % \KH{I must be missing something - Is self-attention of audio standard in transformer encoder?}
While the cross-attention output is passed to the next layer as alignment features, the audio self-attention output is passed to the next layer as audio features. In such a way, attention operations are synchronized according to the audio time, and delays from alignment side and audio side do not add up. 
Specifically, in the schematic diagram of Figure~\ref{fig:streaming_alignrefine}, we have $L=3$ self-attention operations for both alignments and audio, but the effective depth of the architecture $L+1=4$ for determining model delay,\footnote{We could remove audio self-attention at the bottom to make the effective depth $L$ instead of $L+1$, though we have not in this paper.} versus $2L$ if they are stacked as in Figure~\ref{fig:offline_alignrefine}.
% \TS{again refer to the figure at the start of the para not the end to help guide the reader.}

\subsection{Discriminative training}
\label{sec:discriminative}
\vspace*{-.5ex}

It is known in the literature that, after initial training of an ASR model with the MLE criterion, finetuning it with a loss closer to the final evaluation criterion (i.e., WER) often yields further accuracy gain~\cite{vesely2013sequence, shannon2017optimizing, prabhavalkar2018}.
In this work, we perform discriminative training of Align-Refine using the minimum word error rate (MWER) criterion.
For an input utterance $X$, we forward the Align-Refine model to perform $S'$ refinement steps, with $S'$ potentially different from the $S$ used in initial MLE training. And at the end of $S'$ refinement steps, we perform CTC beam search instead of greedy decoding, to output $K>1$ hypotheses denoted as $\hat{y}_1, \dots, \hat{y}_K$. We then compute the log-probabilities of the hypotheses under our model, defined as 
\begin{align*}
    \log P(\hat{y}_k|X) = \frac{1}{S'} \sum\nolimits_{i=1}^{S'} \log P_{ctc}^i (\hat{y}_k|X), \quad k=1,\dots,K. 
\end{align*}
We make the approximation that probability over possible label sequences concentrate in the top-$K$ space, to compute
\begin{align*}
    P_k = \frac{P(\hat{y}_k|X)}{\sum_{y'} P(y'|X)} \approx \frac{P(\hat{y}_k|X)}{\sum_{k=1}^K P(\hat{y}_k|X)},
\end{align*}
and subsequently the MWER loss 
\begin{align*}
    \ell_{\text{MWER}}^{S'} \left( X, y, \{\hat{y}_1, \cdots, \hat{y}_K\} \right) := \sum\nolimits_{k=1}^K P_k \cdot \text{NWE}(\hat{y}_i, y)
\end{align*}
where $\text{NWE}(\hat{y}_i, y)$ measures the number of word errors between hypothesis $\hat{y}_i$ and ground truth $y$. We use superscript $S'$ to signify that hypotheses are obtained after $S'$ refinement steps.
As is common in the literature, for better learning stability, we minimize a composite loss
\begin{align*}
\ell_{\text{MWER}}^{S'} \left( X, y, \{\hat{y}_1, \cdots, \hat{y}_K\} \right) + \gamma \cdot \ell_{\text{MLE}}^{S'} (X, y)
\end{align*}
over the training set in the discriminative training phase, and we fix $\gamma=0.005$ in this work.
Note that while we use beam search in generating multiple hypotheses for MWER training, we still perform greedy decoding for final inference.

\section{Experiments on voice search datasets}
\label{sec:expts-vs}
\vspace*{-.5ex}

In this section, we follow the experimental setup of~\cite{wang2022alignrefine} on voice search datasets so that our results are directly comparable. 
% We refer the readers to Sections~4.1 and~4.2 of~\cite{wang2022alignrefine} for details of datasets and input/output specifications. \TS{you should detail a bit more the train/test sets used rather then refer to previous papers.}
The training set includes anonymized and hand-transcribed multi-domain audio data% covering the search, farfield, telephony and YouTube domains
~\cite{narayanan2019recognizing} and has gone through multi-condition training (MTR~\cite{kim2017generation}) and random 8kHz down-sampling~\cite{li2012improving} as %data
augmentation.
We use a development set of 12K anonymized and hand-transcribed utterances that are representative of Google’s Voice Search traffic, with an average duration of 5.5 seconds; this set is denoted by VS.
Final evaluation is performed on five test sets. The first set is the side-by-side losses test (SXS), containing 1K utterances where the quality of the E2E model transcription has more errors than a state-of-the-art conventional model~\cite{Golan16}. 
The rest four are TTS generated test sets containing rare proper nouns (RPN) which appear less than 5 times in the training set. These sets cover the Maps, News, Play, and QSession domains and are denoted RPN-M, RPN-N, RPN-P, and RPN-Q respectively, each containing 10K utterances.

\subsection{Cascaded RNN-T as first-pass model}
\vspace*{-.5ex}
% \TS{yuou refer to 1st-pass and then discuss causal and non-causal, which is a bit confusing to the reader. change the section title into Cascaded and maybe highlight its causal 1st-pass, non-causal 2nd-pass.}
% Models inputs are 128-dimensional log-Mel features, computed with a 32ms window and shifted every 10ms, followed by frame stacking and subsampling~\cite{narayanan2019recognizing}. 
% Specaugment~\cite{park2019specaugment} is used in the same manner as described in~\cite{park2020specaugment}. Both the first-pass and the deliberation models have a output vocabulary of 4,096 wordpiece units, including the end-of-sentence token.

The first-pass model for generating the initial hypothesis is a cascaded model designed for on-device usage~\cite{sainath2022cascaded}, consisting of two conformer encoders and a embedding-based decoder. The first encoder is causal and is used by the decoder to generate partial results in streaming mode, while the second encoder sits on top of the causal encoder with a right-context of 0.9s and is used for generating more accurate final hypotheses. % Modeling strategies for this model are described in~\cite{sainath2022cascaded}.
% \TS{this is perhaps a lot of detail that can be summarized into fewer sentences, this woudl save space to put back the para on test sets.}
Note that~\cite{wang2022alignrefine} used the causal pass of this model, with a total of 56M weight parameters, to generate initial hypotheses. While we mostly focus on the same setup as it better simulates the streaming usage of Align-Refine, we also provide results for deliberating the non-causal pass which has 155M weight parameters. We freeze the first-pass model for deliberation training. 
The MLE training phase takes 800K steps, with a global batch size of 4096 utterances. % and a transformer learning rate schedule, 
In the MWER training phase, batch size is reduced to 2048 and we evaluate models at 25K steps.

\begin{table}[t]
    \caption{VS WERs (\%) of Align-Refine for different amounts of model delays, with up to three refinement steps during inference. The first pass model is causal RNN-T of 56M weight parameters, and has a WER of 7.8\% on VS.}
    \label{tab:inhouse-dev}
    \vspace*{-1.5ex}
    \centering
    \begin{tabular}{|@{\hspace{0.04\linewidth}}c@{\hspace{0.04\linewidth}}|>{\centering\arraybackslash}p{0.1\linewidth}|>{\centering\arraybackslash}p{0.1\linewidth}|>{\centering\arraybackslash}p{0.1\linewidth}|}
    \hline
        per-step model & \multicolumn{3}{|c|}{Inference refinement step} \\ \cline{2-4}
        delay (secs) & 1 & 2 & 3\\
        \hline\hline
        \multicolumn{4}{|c|}{Alignment self-attention only} \\ \hline
        0.00 & 7.6 & 7.5 & 7.5 \\
        0.36 & 7.3 & 7.2 & 7.1 \\
        0.72 & 6.9 & 6.6 & 6.6 \\
        1.80 & 6.5 & 6.2 & 6.1 \\
        \hline
        \hline
        \multicolumn{4}{|c|}{+ Audio self-attention} \\ \hline
        0.00 & 7.5 & 7.3 & 7.3 \\
        0.42 & 6.6 & 6.4 & 6.3 \\
        0.84 & 6.3 & 6.0 & 6.0 \\
        2.10 & 6.1 & 5.7 & 5.7 \\
        \hline       
        \hline
        \multicolumn{4}{|c|}{+ Continue with discriminative training} \\ \hline
        0.00 & 7.3 & 7.1 & 7.1 \\
        0.42 & 6.5 & 6.2 & 6.2 \\
        0.84 & 6.2 & 5.9 & 5.9 \\
        2.10 & 5.9 & 5.5 & 5.5 \\
        \hline        
    \end{tabular}
    \vspace*{-1ex}
\end{table}

\subsection{Streaming Align-Refine for causal RNN-T}
\label{sec:ar-causal}
\vspace*{-.5ex}

We use a model architecture similar to that of~\cite{wang2022alignrefine} for Align-Refine: the transformer decoder consists of $L=6$ transformer layers with attention dimension 640 and 8 attention heads, and we perform $S=3$ refinement steps during MLE training. We achieve different amount of model delay by varying the number of right-context frames $C$ at each layer, and thus the model delay per refinement step is $L\times C \times f$ where $f$ is the decoder frame size ($60ms$ in this section) without audio self-attention, and $(L+1)\times C \times f$ with audio self-attention.

\begin{table*}[t]
    \caption{Test WERs (\%) on voice search datasets %Align-Refine performs greedy decoding, others perform beam search with a beam size of 4 (RNN-T first-pass) or 8 (the rest). 
    for causal first-pass (top section) and non-causal first-pass (bottom section).}
    \label{tab:inhouse-test}
    \vspace*{-1.5ex}
    \centering
    \begin{tabular}{|@{\hspace{0.01\linewidth}}c@{\hspace{0.01\linewidth}}|@{\hspace{0.01\linewidth}}c@{\hspace{0.01\linewidth}}|@{\hspace{0.01\linewidth}}c@{\hspace{0.01\linewidth}}|r@{\hspace{0.01\linewidth}}||@{\hspace{0.01\linewidth}}r@{\hspace{0.01\linewidth}}|@{\hspace{0.01\linewidth}}r@{\hspace{0.01\linewidth}}|@{\hspace{0.01\linewidth}}r@{\hspace{0.01\linewidth}}|@{\hspace{0.01\linewidth}}r@{\hspace{0.01\linewidth}}|@{\hspace{0.01\linewidth}}r@{\hspace{0.01\linewidth}}|}
    \hline
        \multirow{2}{*}{Method} & Additional & Total model &  \multicolumn{6}{c|}{WERs (\%)} \\ \cline{4-9}
        & weights & delay (secs)  & VS & SXS & RPNM & RPNN & RPNP & RPNQ \\ \hline \hline
        Causal RNN-T & 0 & 0.00 & 7.8 & 37.5 & 16.6 & 11.4 & 40.9 & 25.6 \\
        Attention seq2seq Delib.~\cite{hu2021transformer} & 48M & $\infty$ & 6.0 & 34.3 & \bf 13.8 & 10.2 & \bf 36.2 & 22.2 \\
        Offline Align-Refine~\cite{wang2022alignrefine} & 55M & $\infty$ & 5.7 & 32.0 & 14.6 & 10.0 & 38.3 & 23.5 \\
         
        Streaming Align-Refine & 70M 
        & 0.00 & 7.1 & 35.0 & 15.8 & 11.6 & 39.7 & 24.3 \\
        & & 0.84 & 6.2 & 32.6 & 15.0 & 10.5 & 38.5 & 22.7 \\
        & & 1.68 & 5.9 & 31.1 & 14.5 & 10.2 & 38.3 & 22.3 \\
        & & 4.20 & \bf 5.5 & \bf 30.1 & 14.1 & \bf 9.9 & 37.4 & \bf 21.8 \\
        \hline \hline
        Non-causal RNN-T & 99M  & 0.90 & 5.2 & 27.6 & 12.9 & 9.0 & 37.8 & 20.2 \\
        Attention seq2seq Delib.~\cite{hu2021transformer} & 149M &  $\infty$ & 4.9 & 25.2 & 12.4 & 7.4 & 34.1 & 18.8 \\
        Stream Align-Refine & 169M  & 5.10 & 4.9 & 27.3 & 12.8 & 8.8 & 37.2 & 19.9 \\
        \hline
    \end{tabular}
        \vspace*{-1ex}
\end{table*}

We set the beam size to 4 for the first-pass causal RNN-T model during both training and inference. %; the first-pass has a WER of 7.8\% on VS. 
We provide the WERs of Align-Refine using only alignment self-attention (with time-aligned cross-attention performed on causal encoder outputs) in the top section of Table~\ref{tab:inhouse-dev}. While Align-Refine barely improves over the first pass when not using any right context (i.e., 0.0 sec of model delay), the accuracy improves sharply as the amount of right context increases.
We then include audio self-attention in the model architecture, which leads to larger model size but not much increase in delay, and results of these models are given in the middle section of Table~\ref{tab:inhouse-dev}. Observe that audio self-attention consistently reduces WERs for all model delays, implying that right context in the audio modality is complementary to right context in the text modality.

We then finetune the models with audio self-attention, using the discriminative training procedure discussed in Sec~\ref{sec:discriminative}. Empirically, %at the model delay of 1.80 secs per step, 
we observe no benefit by using more than $S'=1$ refinement steps in $\ell_{\text{MWER}}$, and we stick to this configuration here. % We hypothesize that this is due to the fact that the first refinement step always yields the largest WER reduction from first pass RNN-T, and an improved step $1$ helps with the accuracy of future steps. 
Results of discriminatively trained models are presented in the bottom section of Table~\ref{tab:inhouse-dev}. These models  consistently outperform their MLE-trained counterparts, across all delays.

As noted by~\cite{wang2022alignrefine} for offline Align-Refine, significant WER gains are achieved in the first two refinement steps during inference. Interestingly, we observe a new type of trade-off here: it may be as accurate to run a model of smaller delay for two steps, than to run a model of larger delay for a single step. For example, in the bottom section of Table~\ref{tab:inhouse-dev}, running the discriminatively trained model with 0.42s per-step delay for 2 steps leads to a WER of 6.2\%, the same as running the model with 0.84s per-step delay for 1 step. Similarly, running the model with 0.84s per-step delay for 2 steps leads to a WER of 5.9\%, which is the same as running the model with 2.1s per-step delay, but the former as a smaller total delay of 1.68s.

\subsection{Align-Refine for non-causal RNN-T}
% \TS{i think you mena non-causal 2nd-pass}
\label{sec:ar-non-causal}
\vspace*{-1.0ex}

Given the large WER improvements for deliberating a small first-pass, a natural question is whether our method is still useful when the initial hypothesis is generated by a stronger model. To answer this question, we apply streaming Align-Refine with 2.1s per-step delay to the non-causal RNN-T of 155M parameters (causal encoder + non-causal encoder + RNN-T decoder) and 0.9s model delay.
We feed the causal encoder output to Align-Refine, which has its own audio self-attention. This model improves the first-pass WER of 5.2\% to 5.0\% in one step, and to 4.9\% in another step, without discriminative training (which did not help further).

\subsection{Comparisons with other methods}
\label{sec:final_results}
\vspace*{-1.5ex}

We provide test sets WERs of our models with audio self-attention in Table~\ref{tab:inhouse-test}, along with comparisons with a few models. We take the best streaming %Align-Refine models 
from Sec~\ref{sec:ar-causal} and Sec~\ref{sec:ar-non-causal} and evaluate them with 2 refinement steps. For each method, we provide the amount of additional weight parameters on top of the 56M causal RNN-T, as well as the total model delay.
In the case of causal first-pass (Table~\ref{tab:inhouse-test} top section), we compare with the best offline model from~\cite{wang2022alignrefine}, as well as an attention-based seq2seq deliberation model similar to that of~\cite{hu2021transformer}.
Observe that with sufficient right context (between 1.68s to 4.2s delay), streaming Align-Refine performs as well as the offline counterpart and outperforms prior methods on most test sets with 4.2s delay. 
% \KH{It would be great if you can label each row and reference them during comparison. I think you have quite a bit of results here which is good but I was trying to look for them sometimes.}

For the non-causal first-pass (Table~\ref{tab:inhouse-test} bottom section), we compare again with%the method of
~\cite{hu2021transformer}, where the attention-based decoder performs beam search with beam size 8. Streaming Align-Refine performs similarly to this model on VS (mostly containing frequent words) with a simpler inference procedure, while attention-based deliberation is more accurate on rare words.
% While the total model delay of our model (0.9s + 1.8s $\times$ 2) seems large in the streaming mode, we can practically use the model in offline mode after non-causal RNN-T finishes decoding and endpointing. \TS{this statement is a bit handwavy and might confuse the reader, perhaps clarify better.} With careful batching, we can save on memory usage by leveraging limited-context attention, as opposed to running full-context attention with a offline model.

\section{Experiments on Librispeech}
\label{sec:librispeech}
\vspace*{-.5ex}

We perform experiments on Librispeech~\cite{panayotov2015librispeech} to compare with existing non-AR methods.
The first-pass is a 122M causal RNN-T with conformer encoder and embedding decoder, and uses a beam size of 8 for inference. The token set contains 1024 wordpieces. We use the streaming Align-Refine architecture found on voice search (70M parameters, including audio self-attention) without further tuning. 
We have trained four models of different per-step model delays: 0.84s, 2.1s, 4.2s, and 21.0s, with the last setup simulating offline Align-Refine. We do not find discriminative training to be helpful in this setup, and report WERs from MLE training. 
We compare our results with those of offline non-AR methods~\cite{chan20,chi2020align}, as shown in Table~\ref{tab:librispeech}. % \TS{you might want to pint out this is state of the art of Librispeech for non-AR methods??}
Note that our first-pass RNN-T already has WERs similar to the final WERs of prior work. Streaming Align-Refine consistently improves over the strong first-pass, and its accuracy steadily increases with model delay. 
% \TS{this para is out of place relative to the rest of the paper. Also isnt this more important for VS compared to Librispeech. you might want to add section 3.5 and add this in there. if you can compare to delib or non-causal RNN-T RTF thats even better, it would fit nicely with Table 2.}
We measure the inference speed of methods with a single Intel Xeon CPU (@2.20GHz). The real-time-factor (RFT) of the first-pass is 0.123, while the RTF of streaming Align-Refine is 0.045 per refinement step.

\begin{table}[t]
    \caption{Librispeech WERs (\%) of non-AR methods.}
    \label{tab:librispeech}
    \vspace*{-1.5ex}
    \centering
    \begin{tabular}{|@{\hspace{0.04\linewidth}}c@{\hspace{0.04\linewidth}}|>{\centering\arraybackslash}p{0.14\linewidth}|>{\centering\arraybackslash}p{0.14\linewidth}|>{\centering\arraybackslash}p{0.14\linewidth}|>{\centering\arraybackslash}p{0.14\linewidth}|}
    \hline
        Method & dev\_clean & dev\_other & test\_clean & test\_other \\
        \hline
        First-pass & 3.6 & 9.5 & 4.0 & 9.1 \\ \hline
        \multicolumn{5}{|l|}{Streaming Align-Refine per-step delay and WERs @step 1/2} \\ \hline
        0.84s & 3.5/3.4 & 9.1/8.9 & 3.7/3.6 & 8.8/8.6 \\
        2.10s & 3.2/3.1 & 8.6/8.3 & 3.4/3.3 & 8.5/8.2 \\
        4.20s & 3.1/3.0 & 8.4/8.0 & 3.3/3.2 & 8.3/8.0 \\
        21.0s & \bf 2.9/2.8 & \bf 8.0/7.7 & \bf 3.2/3.0 & \bf 7.8/7.6 \\
        \hline      
        \multicolumn{3}{|l|}{Imputer~\cite{chan20}} & 4.0 & \hspace{-1.5ex} 11.1 \\
        \multicolumn{3}{|l|}{Offline CTC +  Align-Refine~\cite{chi2020align}} & 3.6 & 9.0 \\
        \hline
    \end{tabular}
    \vspace*{-1ex}
\end{table}

\section{Conclusions}
\label{sec:conclusions}
\vspace*{-.5ex}

We have proposed a streaming non-autoregressive decoding method for second-pass deliberation. Our method improves WERs of both small and large streaming RNN-T models with controllable model delay, and benefits from discriminative training when the first-pass has small capacity. % \KH{somewhere in the paper you may want to explicitly mention that the proposed streaming model does not introduce additional delay. Would be great to discuss a little about latency because the highlight here is streaming.} The newly proposed transformer architecture performs efficient integration of text and acoustic features, and enables trade-off between accuracy and model delay. 
As a future direction, we would like to incorporate large amount of unpaired data into our model training~\cite{deng2022improving,hu2022improving}, % \KH{maybe cite our work on using BERT as text encoder?}
to better capture label dependency and improve on rare word recognition.

% References should be produced using the bibtex program from suitable
% BiBTeX files (here: strings, refs, manuals). The IEEEbib.bst bibliography
% style file from IEEE produces unsorted bibliography list.
% -------------------------------------------------------------------------
\bibliographystyle{IEEEtran}
\pagebreak
\bibliography{refs}

\end{document}